\documentclass[10pt, a4paper]{article}
\usepackage{lrec2022} 
\usepackage{multibib}
\newcites{languageresource}{Language Resources}
\usepackage{graphicx}
\usepackage{tabularx}
\usepackage{soul}
\usepackage{titlesec}
\titleformat{\section}{\normalfont\large\bfseries\center}{\thesection.}{1em}{}
\titleformat{\subsection}{\normalfont\SmallTitleFont\bfseries\raggedright}{\thesubsection.}{1em}{}
\titleformat{\subsubsection}{\normalfont\normalsize\bfseries\raggedright}{\thesubsubsection.}{1em}{}
\renewcommand\thesection{\arabic{section}}
\renewcommand\thesubsection{\thesection.\arabic{subsection}}
\renewcommand\thesubsubsection{\thesubsection.\arabic{subsubsection}}
\usepackage{booktabs}
\usepackage{epstopdf}
\usepackage[T1]{fontenc}
\usepackage[utf8]{inputenc}
\usepackage[icelandic]{babel}

\usepackage{hyperref}
\usepackage{xstring}

\usepackage{color}
\usepackage{ulem}

\title{Building an Icelandic Entity Linking Corpus}

\name{Steinunn Rut Friðriksdóttir\textsuperscript{1}, Valdimar Ágúst Eggertsson\textsuperscript{3}, Benedikt Geir Jóhannesson\textsuperscript{2}, \\
\large \textbf{Hjalti Daníelsson\textsuperscript{3}, Hrafn Loftsson\textsuperscript{2}, Hafsteinn Einarsson\textsuperscript{1}}} 

\address{\textsuperscript{1}University of Iceland, \textsuperscript{2}Reykjavík University, \textsuperscript{3}Quick Lookup \\
         Reykjavík, Iceland \\
         \textsuperscript{1}\{srf2, hafsteinne\}@hi.is, \textsuperscript{2}\{benediktj20, hrafn\}@ru.is, \textsuperscript{3}\{valdimar, hjalti\}@snjallgogn.is}

\abstract{
In this paper, we present the first Entity Linking corpus for Icelandic. We describe our approach of using a multilingual entity linking model (mGENRE) in combination with Wikipedia API Search (WAPIS) to label our data and compare it to an approach using WAPIS only. We find that our combined method reaches 53.9\% coverage on our corpus, compared to 30.9\% using only WAPIS. We analyze our results and explain the value of using a multilingual system when working with Icelandic. Additionally, we analyze the data that remain unlabeled, identify patterns and discuss why they may be more difficult to annotate. 
 \\ \newline  }

\begin{document}
\renewcommand{\abstractname}{Abstract}
\renewcommand{\figurename}{Figure}
\renewcommand{\tablename}{Table}

\maketitleabstract

\section{Introduction}
In recent years, Natural Language Processing (NLP) has progressed rapidly. New solutions in NLP have led to more effective human-computer interaction and easier access to on-demand knowledge~\cite{balog2018entity}. Before this development, the analysis of unstructured data posed a severe challenge if attempted without significant human input and domain knowledge. As a result, there has been growing interest in developing methods to work efficiently with unstructured data.

Information Extraction (IE) is the process of automatically retrieving structured information from unstructured, machine-readable sources. Such structured information can, for example, refer to Named Entities (NEs) found in any given text, the relationship between different entities, and the attributes that describe them. IE enables much deeper and more complex queries for such information from a far wider variety of sources~\cite{sarawagi2008information}.

Prior work within the field of IE has focused on methods to recognize entities in text, which is known as Named Entity Recognition (NER). NER methods aim to automatically recognize NEs in text and assign them to appropriate predefined categories, like \texttt{Person}, \texttt{Organization} and \texttt{Location}. However, mentions can often be ambiguous and refer to different real-world entities depending on their context. For instance, a NER system does not differentiate between \textit{Barack} and \textit{Michelle} when both are referred to as \textit{Obama}. This example demonstrates the need for Entity Linking (EL) and Named Entity Disambiguation (NED)\footnote{It should be noted that these terms are often used interchangeably. Some refer to the entire process as Entity Linking.}. After the NER task, the EL system looks the entities up in a Knowledge Base (KB), either a first-party one that has been created for the EL task, or a third-party one, such as Wikidata, and links their mentions to records in the KB. If multiple records are found for a given mention, the NED system performs disambiguation to select the correct entity based on the given context. The EL task is complete when the NEs have been disambiguated and correctly linked to the KB. 

Building systems for NED and EL using state-of-the-art methods requires training data, i.e. a sufficiently large text corpus where mentions have been linked to correct entities in a KB. Building such a corpus and a KB can demand a significant effort, since it requires the labelling of mentions, the creation of entities in a KB, and the task of linking mentions to their corresponding records. The effort required can be a barrier to developing good NED and EL systems. Therefore, it is essential to develop methods that reduce the work required to create the training data.

In this paper, we present a method we used to efficiently build the first Icelandic corpus where entities are linked to corresponding records in a KB\footnote{ Our corpus has been made publicly available on CLARIN-IS: \url{https://repository.clarin.is/repository/xmlui/handle/20.500.12537/168}}. The underlying data is based on texts from diverse sources and is essential for any type of NED work in Icelandic as international corpora and KBs do not successfully reflect local entities and country-specific information. We believe that the method presented in this paper can be beneficial to those who want to bootstrap EL corpora for other lower-resource languages where entities are linked to the Wikidata KB.

The rest of this paper is structured as follows. In Section~\ref{related}, we discuss previous work in the field of EL, particularly in relation to multilingual systems, and explain their significance to our work. In Section~\ref{corpus}, we present our corpus and the methodology used for its compilation. Section \ref{analysis} analyses our corpus as well as the performance of the methods used for its creation. We analyze the data that remain unlabeled in Section \ref{remaining} and explain which factors might cause difficulties in the labeling process.

\section{Related work} 
\label{related}

Most publicly available EL corpora use Wikidata as a KB (e.g. \newcite{hoffart2011robust}, \newcite{nuzzolese2015open}, and \newcite{minard2016meantime}). The focus has been on text diversity in recent years, since training on professionally curated corpora, such as news articles, may not generalize well to other domains, such as text from social media. For example, \newcite{eshel2017named} compiled their EL corpus by crawling the web searching for links to Wikipedia. This way, they constructed the WikilinksNED corpus, which consists of Wikipedia hyperlinks and their surrounding context, using page IDs as unique identifiers for entities. As another example, \newcite{botzer2021reddit} presented an EL corpus of 17,316 entities collected from Reddit, manually annotated by Mechanical Turk workers who matched mentions to Wikipedia links.  

Most work in the field of EL has focused on English, but multilingual EL has received increased attention in the last 10 years or so. Originally, most multilingual EL systems linked mentions in a specific language or languages to a KB in another, higher-resource language such as English (e.g. \newcite{mcnamee2011cross}, \newcite{mayfield2011building}, \newcite{ji2015overview}). In contrast, \newcite{botha2020entity} proposed a method where language-specific mentions are linked to a language-agnostic Wikipedia-based KB. Their model covers over 100 languages and 20 million entities, making the EL process more inherently multilingual. Following their lead, \newcite{de2021multilingual} presented a sequence-to-sequence system, mGENRE, for multilingual EL, which is the system we use in our corpus generation process. 

mGENRE is a multilingual version of the GENRE model~\cite{de2020autoregressive}, trained on large corpora in 125 languages and covering a range of $\sim$730M Wikipedia hyperlinks in 105 languages. The model is an auto-encoder based on the BART architecture~\cite{lewis-etal-2020-bart}. For a given input text, mGENRE generates language IDs and entity names that, in combination, uniquely identify records in the Wikidata KB. Importantly, mGENRE does not require an external KB at runtime since information about entities is contained in its trained network parameters. Furthermore, by maintaining entity names in as many languages as possible, mGENRE is able to exploit connections between languages along with interactions between the source mention context and the target entity name. During inference, beam search is used to determine the probability scores for candidates. The scores for different languages are marginalized in order to score entities~\cite{de2021multilingual}.
It is worth noting that mGENRE performs EL and NED simultaneously.

In this work, mGENRE is used to suggest records in Wikidata to speed up the EL labeling process in an Icelandic corpus. We would like to emphasize that, as Icelandic is one of the 105 languages covered by mGENRE, the model can be applied directly to our data. Thus, a replication of this study, for the other languages that mGENRE covers, should be eminently feasible. The corpus is based on an annotated NER corpus, MIM-GOLD-NER~\citelanguageresource{20.500.12537/140}, which contains around 48,000 NEs~\cite{ingolfsdottir2020named}. The corpus is tagged for eight NE types (\texttt{Person}, \texttt{Location}, \texttt{Organization}, \texttt{Miscellaneous}, \texttt{Date}, \texttt{Time}, \texttt{Money} and \texttt{Percent}) and is in the CoNLL format. We advance MIM-GOLD-NER to the next logical step by building a new corpus, MIM-GOLD-EL, in which NEs of type \texttt{Person}, \texttt{Location}, \texttt{Organization}, and \texttt{Miscellaneous} from MIM-GOLD-NER are linked to unique entities in the Wikidata KB.  

Icelandic Language Technology (LT) has made notable advances recently, not least in relation to a national funding program aimed at creating the necessary resources for further advancement in the field \cite{nikulasdottir2020language,nikulasdottir2021buffet}. This has resulted in the publication of multiple LT tools and resources during the last three years or so, which has brought Icelandic into a medium-resource language tier. We benefit greatly from the fact that the MIM-GOLD-NER corpus has already been published, making our work significantly easier. While Icelandic does not technically qualify as a low-resource language\footnote{A low-resource language is a language for which few on-line resources exist or for which few computational data exist \cite{cieri-etal-2016-selection}.} anymore, we still believe that our work can be considered beneficial for languages in need of EL data, particularly those that are covered by mGENRE. As mGENRE builds on Wikipedia as a foundation, we note that efforts to build Wikipedia in a given language can lead to  downstream benefits such as better EL in models such as mGENRE.

\begin{figure}[!t]
\begin{center}
\includegraphics[width=0.7\columnwidth]{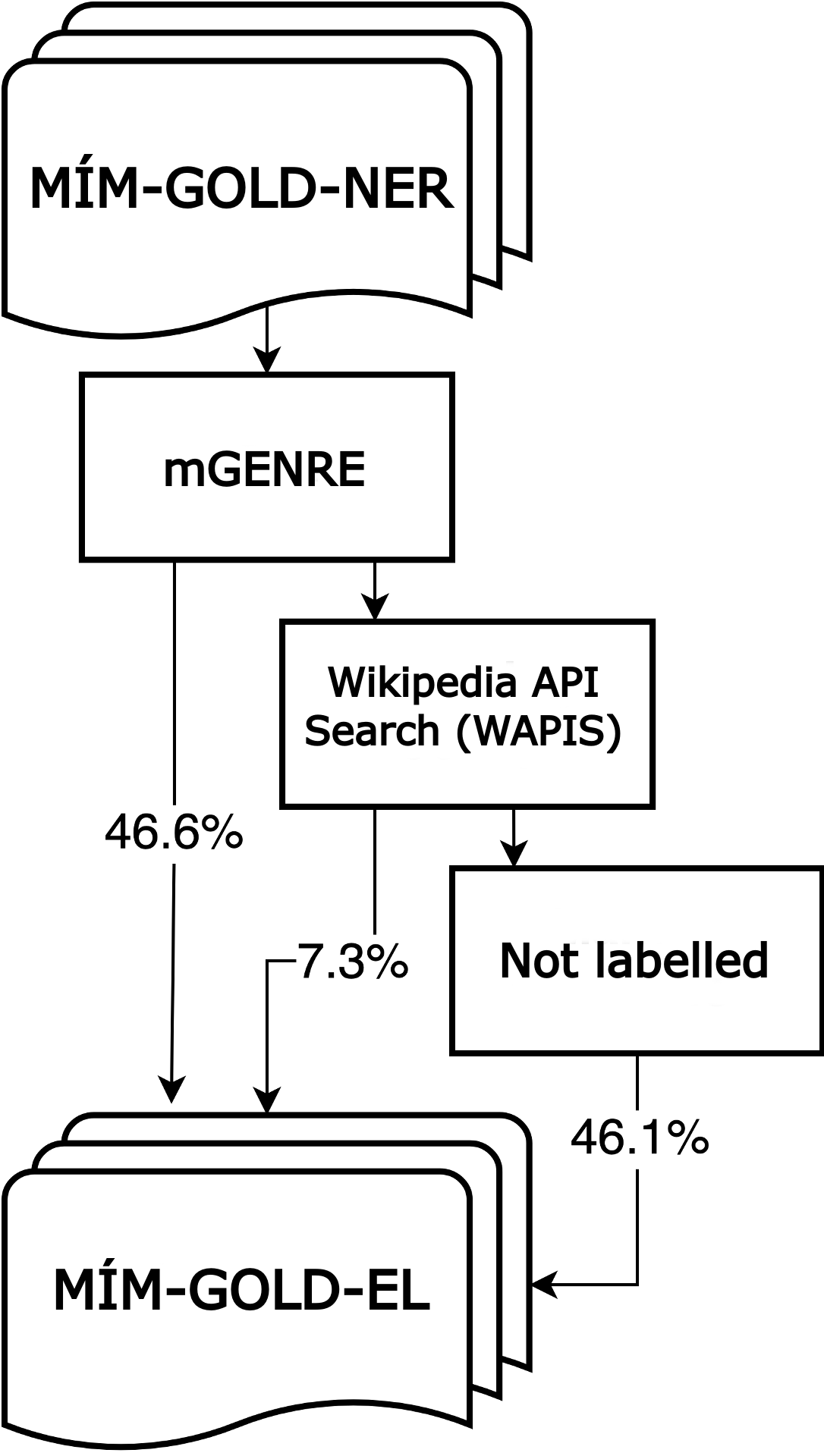}
\caption{Our corpus compilation process with proportions of mentions retrieved for each step, i.e.\ 46.6\% of all mentions are labeled by mGENRE and subsequently 7.3\% of all mentions are labeled using the WAPIS. Of all mentions, 46.1\% were not identified and were marked as \texttt{unlabeled} label in MIM-GOLD-EL.}
\label{fig.1}
\end{center}
\end{figure}

\section{Corpus compilation}
\label{corpus}
In order to create our corpus, MIM-GOLD-EL, we started by preprocessing MIM-GOLD-NER in accordance with the input format required for mGENRE. For each entity, mGENRE generates a set of identifiers (IDs) that consist of pairs of the language in question and the name of the entity in said language. Each Wikidata item has a set of Wikipedia pages in multiple languages linked to it, thus the task of mGENRE is to uniquely identify the entities using these IDs. We ran the data through mGENRE, which proposed Wikidata IDs for the retrieved entities and set them as labels for the mentions. We checked our results manually, accepting or rejecting each of the model's predictions. This resulted in 46.6\%, of the 39,793 mentions examined, being accepted.

Subsequently, we ran all mentions through a separate process, which we refer to as Wikipedia API Search (WAPIS). In this process, we used the text of each mention in a search query run on the Wikipedia API, specifically the Icelandic and the English wikis. The query is the equivalent of entering the full text of the mention into a Wikipedia search window, and harvesting the suggested links that would appear below in a drop-down list. For each such query, we used the full text of the mention, unaltered (e.g. not lemmatized) and not including any surrounding context in the source texts.

Mentions that had remained unlabeled after the mGENRE round, and had acquired at least one Wikipedia link after the WAPIS, had their set of links manually reviewed. If one of those links was a match for the mention, we marked that link as the correct label, with preference given to Icelandic links whenever possible. Out of the full set of all mentions, only 7.3\% were found by WAPIS but not by mGENRE\footnote{Due to the nature of our approach, we did not study a WAPIS-only process since it would have resulted in a significant amount of additional work. However, we note that  languages not covered by mGENRE but with a sizeable Wikipedia, could benefit from a WAPIS-only approach.}. Mentions that were neither covered by mGENRE nor WAPIS are also a part of our dataset and may be distinguished from the linked mentions. We cover these mentions in the next two sections and discuss why they might remain unlabeled\footnote{Note that a mention is only marked as unlabeled in the released data if both the \textit{suggestion wiki} (WAPIS-only result) and \textit{correct wiki} (mGENRE result) fields are empty, indicating that neither of our methods were successful.}.

Our purpose with WAPIS was twofold: First, to see if we could increase the amount of correct labels for our mentions, and, secondly, to evaluate mGENRE's overall output when compared to that of a simple Wikipedia text search. It turned out that WAPIS did increase the amount of correct labels, and mGENRE outperformed WAPIS. As noted above, mGENRE's results resulted in 46.6\% of mentions being accepted. The subset of those accepted mentions that also had the correct label in the subsequent WAPIS was 23.6\% (these were not manually reviewed, since the correct candidate had already been established). The entirety of mentions confirmed by WAPIS, irrespective of whether they'd been labeled earlier by mGENRE, was 30.9\%, while the total coverage of our combined approaches of mGENRE and WAPIS was 53.9\%. The process is illustrated in Figure \ref{fig.1}. Finally, using these results we conclude that mGENRE's accuracy on labeled entities in our corpus is 86.4\%\footnote{This is calculated by dividing the number of words labeled by mGENRE by the number of words labeled by both methods}.

\begin{figure}[!h]
\begin{center}
\includegraphics[width=\linewidth]{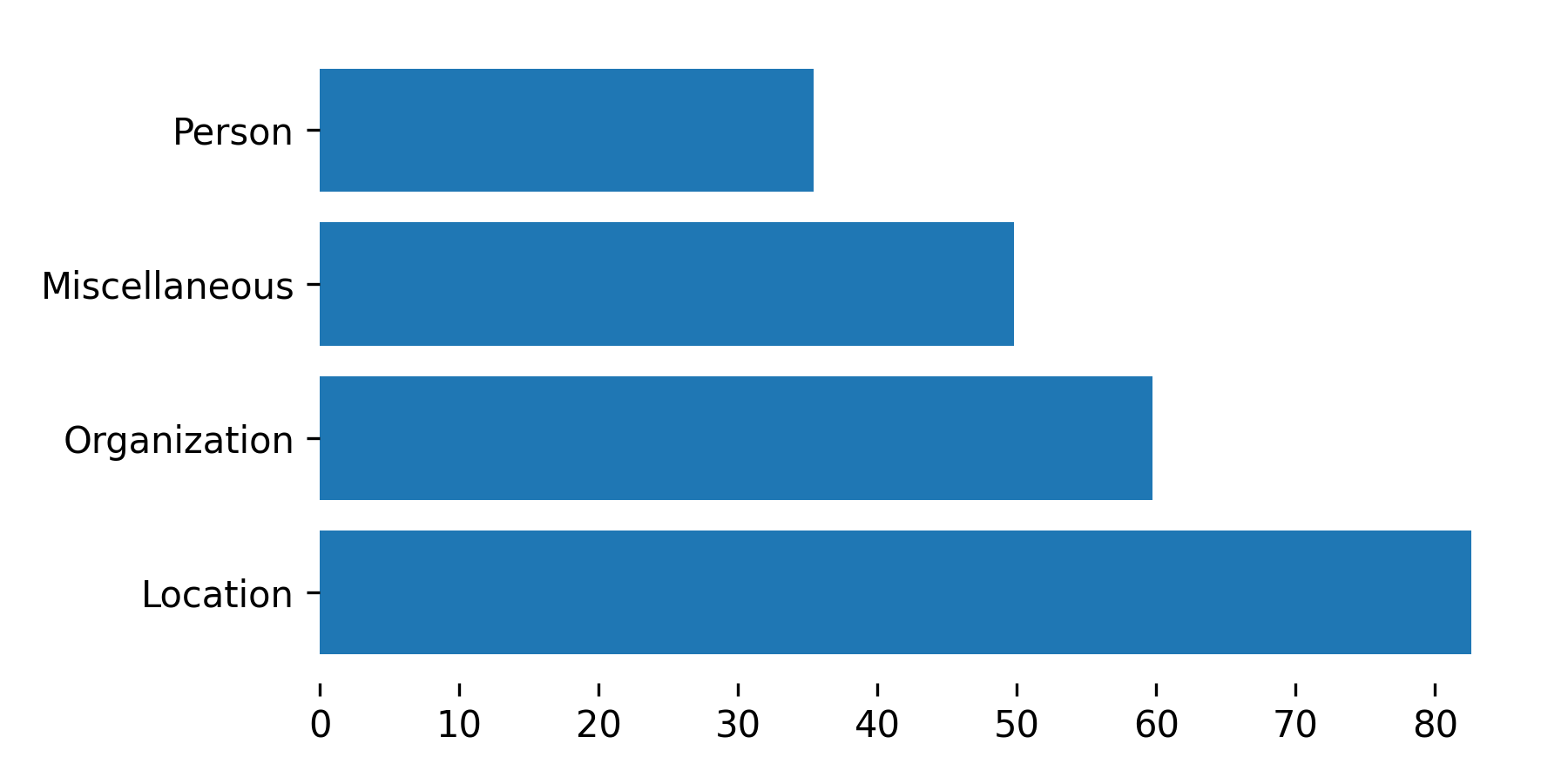}
\caption{Ratio of mentions retrieved by our methods for each NER type.}
\label{fig.4}
\end{center}
\end{figure}

Figure \ref{fig.4}, shows the ratio of mentions covered by our methods for each NER type. The \texttt{Location} type is the most easily retrieved by our methods, followed by \texttt{Organization} and \texttt{Miscellaneous}. The lowest scoring category is that of \texttt{Person}. This is most likely due to how many of the unlabeled mentions refer to people or fictional characters that do not have a corresponding Wikidata entry. On the other hand, locations are generally well documented and infrequently lack their corresponding entries. The \texttt{Miscellaneous} category includes mentions that refer to products, books and movie titles and events.  

The overall process was completed in approximately one month and performed by four annotators. mGENRE was run using a GeForce RTX 2070 SUPER 16GB. No specific computational power is needed for WAPIS. 

\begin{figure*}[!h]
\begin{center}
\includegraphics[width=\linewidth]{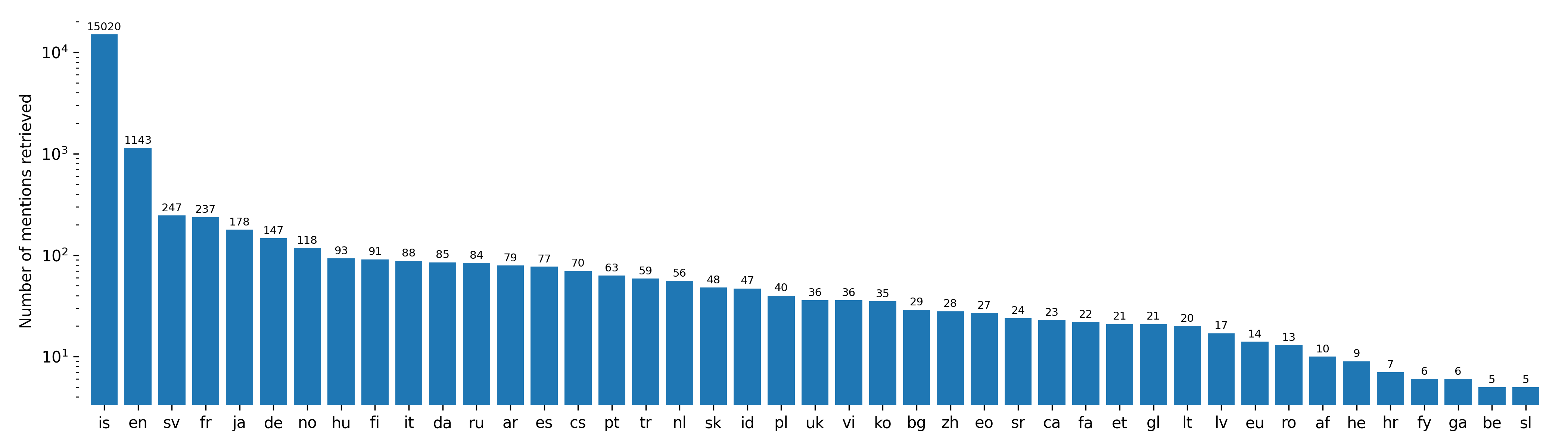}
\caption{Number of mentions retrieved by mGENRE per language. The majority of mentions retrieved were in Icelandic (81.3\%) followed by English (6.2\%) and Swedish (1.3\%). Languages with fewer than five linked records are omitted.}
\label{fig.2}
\end{center}
\end{figure*}

\begin{figure*}[!h]
\begin{center}
\includegraphics[width=\linewidth]{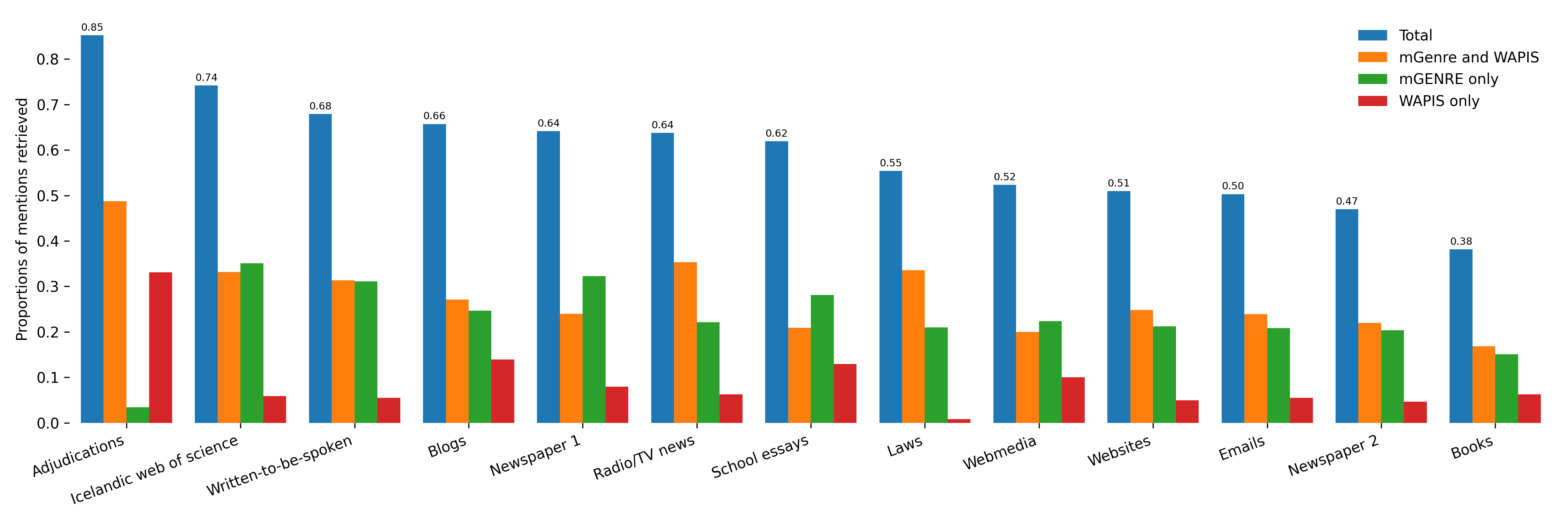}
\caption{Proportions of mentions retrieved per text subcategory in the MIM-GOLD-NER corpus.}
\label{fig.3}
\end{center}
\end{figure*}

\section{Exploratory Corpus Analysis}
\label{analysis}
It is apparent when examining the results that context-aware language models (like mGENRE) provide significant improvements over a plain search query lookup. Using only the WAPIS would have resulted in only a 30.9\% coverage rate. Additionally, there are significant benefits in using multilingual EL methods when working with low-resource languages. Our entities are retrieved from Wikipedia pages in 68 different languages. As shown in Figure~\ref{fig.2}, the most common languages are Icelandic and English which constitute 81.3\% and 6.2\% of the entities retrieved by mGENRE, respectively. Other languages therefore account for 12.8\% of the retrieved entities. While it is apparent that Icelandic makes up the majority of all retrieved entities (75.3\% of the entities retrieved by the WAPIS were also from the Icelandic Wikipedia), it is still safe to assume that our coverage rate would have been significantly lower if we had restricted ourselves to the Icelandic Wikipedia. 

It's worth noting that using a model such as mGENRE might introduce some biases. Looking at Figure~\ref{fig.2}, we see that the most common languages beside Icelandic and English are Swedish (1.3\%), French (1.3\%), Japanese (1\%), German (0.8\%) and Norwegian (0.6\%). Most of these languages are close to Icelandic, both in a cultural and linguistic sense, particularly the Nordic languages. Additionally, at the time of writing, the French edition of Wikipedia has the third largest number of articles published and the German edition has the seventh\footnote{See the \href{https://en.wikipedia.org/wiki/List_of_Wikipedias}{list of Wikipedias} wiki page.}. While it might seem slightly surprising to see Japanese among these languages, it should be noted that most mentions retrieved by the Japanese Wiki were actually Japanese car models. In any case, having all these languages as available resources significantly improves the coverage rate. 

As can be seen in Figure~\ref{fig.3}, the performance of mGENRE and WAPIS is quite varied for different text subcategories in our corpus. The lowest performance was in the books category where only 38.2\% of mentions were linked to their corresponding entities. This is explained by the fact that a lot of the fictional characters who appear in these books do not have Wikidata entries, and by the nature of the way books are written, these types of mentions appear very frequently in the text. It is a bit surprising how high the blog category scores as it includes plenty of mentions that lack proper context for disambiguation and might refer to people that are not public figures. It is also interesting that there is a significant difference in performance for the two newspaper categories, Fréttablaðið at 64.2\% and Morgunblaðið at 46.9\%. Currently, we do not have an explanation for this difference. It is, however, apparent that the highest scoring categories are generally those that have most likely been proofread, the highest scoring of which is the adjudications,\footnote{\label{foot8}It should be noted that the adjudications texts include anonymized names (both for persons and locations) so the coverage rate is skewed by the fact that if the text contains the pseudonym A, we mark it as a correct label if our methods suggest a Wiki entry about the letter A. This category is included because it is a part of the original corpus but needs significant revisions to be useful for EL purposes.} followed by the Icelandic Web of Science, an academic page run by the University of Iceland.   

\begin{figure}[!h]
\begin{center}
\includegraphics[width=\linewidth]{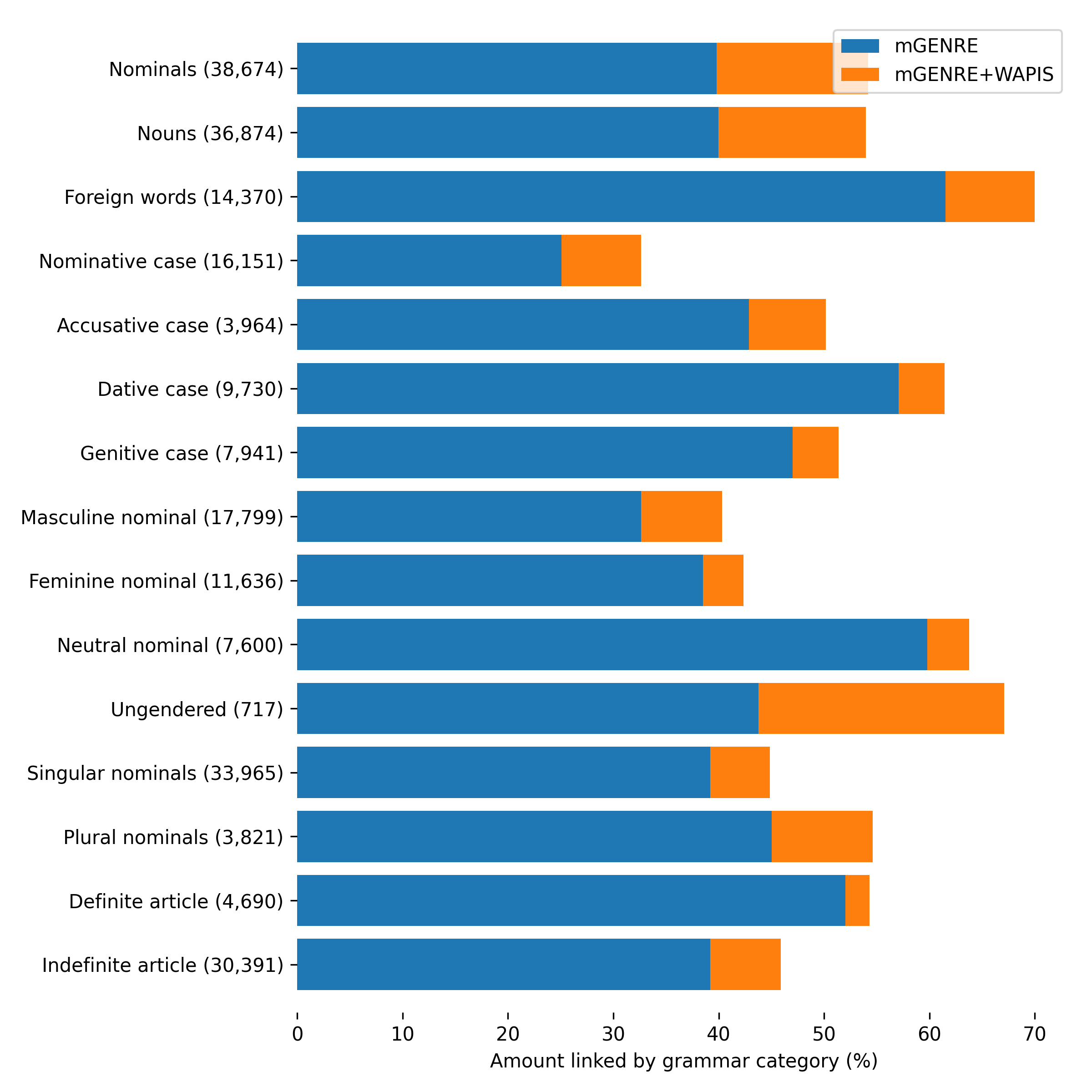}
\caption{Proportions of mentions retrieved per morphological category in the MIM-GOLD-NER corpus. The total number of words for each category is shown within brackets.}
\label{fig.5}
\end{center}
\end{figure}

We analyzed our results in accordance with the fact that Icelandic is a morphologically rich language. In this analysis, each word is treated individually, thus multi-word mentions are treated as multiple words. The total number of words examined was 56,732, out of which 38,674 were nominals (including 36,874 nouns). Foreign words were 14,370. In Figure \ref{fig.5}, the coverage rate of each morphological category can be seen. It is clear that the overall trend is in line with the overall coverage rate of our methods: mGENRE reaches approximately 7\% less coverage than the combined methods (the difference ranges from 2.26\% to 14.4\%, excluding the ungendered nouns where the difference is 23.3\%) and does not appear to struggle with any morphological category in particular. Our analysis thus indicates that mGENRE shows great potential when used for morphologically rich languages.  

\begin{table*}[ht]
\centering
\begin{tabular}{p{0.23\linewidth}p{0.35\linewidth}p{0.35\linewidth}}
\toprule
\textbf{Icelandic} & \textbf{Translation} & \textbf{Probable reason for non-retrieval}\\\midrule
Vesturlandabúans & Inhabitant of the Western countries & Low frequency, oblique case. No specific Wiki entry. \\
Borginni & The City & Oblique case, refers to a diner/club that doesn't exist anymore. No specific Wiki entry. \\
Aðalbygging & The Main Building (refers to the main building of the University of Iceland) & A common moniker, no specific Wiki entry. \\
Jóni Hjaltalín Ólafssyni & A doctor's name & Oblique case, this person does not have a Wiki entry. \\
Mersault & Meursault (French Wine) & The name is misspelled in our data. \\
Bjössi & A name of a fictional character & A common nickname of a person, this book character does not have a Wiki entry. \\
Jón & A person's name (refers to Jón Steinsson, economist) & A common name of a person which, standing by itself, lacks context for disambiguation. \\
Forliti & A person's last name & Stands by itself in a parenthetical reference, lacks context for disambiguation. \\
Svarta kortið & The Black Card (a credit card for students) & Low frequency, no Wiki entry. \\
Swann & A person's last name (refers to Charles Swann, fictional character) & Stands by itself, lacks context for disambiguation. \\
Barnið & An Icelandic translation of the Belgian movie title, L'Enfant & Translated movie titles hardly ever get their own Wiki entries. L'Enfant does not have an entry in the Icelandic Wikipedia. \\
Steinar & A person's name (the text refers to 'Steinar bóndi í Hlíðum', meaning a farmer from a specific farm that has a relatively generic name) & Could refer to multiple people, little chance of disambiguation without more context. \\
Digranesvegi & A street name & Oblique case, no specific Wiki entry. \\
Herdísar L. Storgaard & A nurse's name & Oblique case, this person does not have a specific Wiki entry. \\
Ólafur Gísli Jónsson & A person's name & This person does not have a specific Wiki entry. \\
Miðausturland & The middle of the East side of Iceland & Very specific, yet non-specific location. No Wiki entry. \\
EÖÞ & Abbreviation of an author's name & Stands by itself. Very hard to disambiguate without more context. \\\bottomrule
\end{tabular}
\caption{Examples of mentions not retrieved/labeled by our methods and proposed explanations for the failure.}
\label{tab:examples}
\end{table*}

While our methods have reached a decent coverage rate, there remains a lot of room for improvement as 46.1\% of our data remain unlabeled. Table \ref{tab:examples} shows some examples of the words not retrieved by our methods. We find that relying solely on Wikipedia and on Wikidata as a KB, even with 125 different language versions available, falls significantly short of complete coverage. Some improvements could be made by automatically creating new Wiki entries for the mentions that do not exist in the KBs. This would certainly work for well-known people, fictional characters and locations that coincidentally do not already have their own entry. However, that does not solve everything as can clearly be seen within our blog subcategory. There, it is essentially impossible to disambiguate a lot of the mentions, as they refer to common people and the context of the text might be long forgotten. How to solve the EL task for noisy data such as from social media therefore still remains an open question in this context.   

\section{Analysis of the Remaining Data}
\label{remaining}
As illustrated in Figure~\ref{fig.1}, 46.1\% of the mentions were not covered by the first two steps of the process, mGENRE and WAPIS. Looking at the remaining unlabeled data, we see some patterns. First, some of the mentions are either specific to Icelandic context or infrequent in everyday language. These mentions do not have a corresponding Wikipedia record and therefore cannot be discovered by the first two steps of the process. Second, it is difficult to properly disambiguate last names by themselves, particularly when they are used for reference (e.g. in parenthetical referencing in academic text, mostly found in articles from the Icelandic Web of Science). In most cases, last names will not be retrieved by the WAPIS method either, as it assumes that the mention refers to a first name. Third, abbreviations prove difficult in most cases, except when the entity is more commonly referred to by its abbreviation than its actual name (e.g. NASA). Examples of these types of non-retrieved mentions can be found in Table \ref{tab:examples}.

We manually examined the remaining data in order to gain a better understanding of what type of data specifically causes our methods to fail to return a label. Out of the 18,402 unlabeled entities, 7,236 or 39.3\% refer to a person (most commonly academics, musicians, and athletes) and 2,706 or 14.7\% refer to a fictional character. Institutions and companies make up 12.4\% of the unlabeled data, particularly noticeable of which are churches and public institutions that have abbreviations (e.g. \textit{Landspítalinn} (the largest public hospital in Iceland) often gets referred to as \textit{LSH}). Locations take up 10.1\%, most commonly street names, clubs, restaurants or farms. Among the categories that account for less than 3\% of the unlabeled data are book titles, brands, events, radio and TV shows, nomenclatures and magazine titles. Interestingly enough, there are also 28 mentions of God (particularly when referred to as \textit{the Lord}) that our methods do not cover.   

While examining these categories, we also annotated the data based on specific factors that might impact the model's ability to decipher their meaning. 4,826 or 26.2\% of all mentions refer to a person using only their first name, which is almost invariably done with Icelandic names unless they are appearing for the first time in a given text. On the other hand, 730 or 4\% of the mentions refer to a person only by their last name, and nicknames account for 4.3\%. Other factors that can misdirect our methods from retrieving labels for peoples' names is when there is an insertion between the first and the last name (i.e. \textit{Halldór \textbf{heitinn} Laxness}, `\textit{\textbf{the late} Halldór Laxness}'). We considered several other factors such as abbreviations (4\%), a total lack of context (1.2\%), inexact locations such as \textit{Asian countries} (0.7\%) and Icelandic translations of foreign titles (0.5\%).   

Clearly, there are a lot of components to consider when using automatic methods for entity linking. While the process is made a lot quicker and simpler by the use of language models that greatly reduce the need for manual annotation, they still fall short in many cases where a human annotator might not have any difficulties reaching the correct conclusion. As previously stated, however, a part of the problem is that when relying solely on Wikidata as a KB, we can never reach full coverage, particularly on entities that appear very rarely. For rare entities with unique names, this is not a problem, but once disambiguation is required, the task can become significantly more challenging, even for a person with access to an online search engine. In such cases, it is not clear what the entity should be linked to. One way to approach this problem is to provide some reference, like a web link to a record in an archive that can be considered a good source for that item. Another approach could be to build a larger Wikipedia, possibly via some form of automation. While such approaches may be promising, they present several challenges, especially concerning validation. We emphasize that references to rare entities outside existing knowledge bases is an important unsolved problem and we do not cover it further in this paper.

\section{Ethical Considerations and Broader Impact}
When working with language data, it is important to consider the ethical implications of one’s work. In the context of entity linking, this includes ensuring that the data used to train and test entity linking models is representative of the real-world distribution of entities, and that the entities in the data are linked correctly and accurately. Furthermore, in the case of the Icelandic entity linking dataset, the data was partially collected from public websites such as blogs without the consent of the individuals involved. While the data is public, it is possible that some individuals may not want their data to be used for research purposes.

Entity linking datasets can have a broad impact beyond the immediate research context in which they are used. For example, a dataset of entities linked to Wikipedia pages could be used to improve search results for a given entity, or to generate summaries of entities for a given user. Furthermore, a dataset of entities linked to named entities could be used to improve the accuracy of named entity recognition models, which is an important component of many natural language processing applications.

\section{Conclusion} 
\label{conclusion}

Icelandic is a morphologically rich language with its own sets of challenges when it comes to creating an EL system. No training data has previously existed for this purpose and thus the first milestone in our journey has been the creation of an EL corpus, MIM-GOLD-EL, which is based on an existing NER corpus, MIM-GOLD-NER. We used mGENRE, a sequence-to-sequence EL model, in order to label our corpus and improved our results using Wikipedia API Search. We analyzed our methodology with regards to reducing manual labour and examined how the morphology of Icelandic can be a factor in our results. Furthermore, we presented a detailed analysis on the data not covered by our methods.

Future milestones will include the creation of a comprehensive and open knowledge graph (KG) of Icelandic NEs which is an essential foundation of most EL projects as well as NED related research and development. Additionally, we will adapt a proven, state-of-the-art technology in order to create an Icelandic Entity Linker, the first of its kind. 

\section{Acknowledgments}
This work was funded by the Icelandic Strategic Research and Development Program for Language Technology 2021, grant no. 200075-5301.

\section{Bibliographical References}\label{reference}

\bibliographystyle{lrec2022-bib}
\bibliography{main}

\section{Language Resource References}
\label{lr:ref}
\bibliographystylelanguageresource{lrec2022-bib}
\bibliographylanguageresource{languageresource}

\appendix
\section{Data statement}
This statement follows the schema proposed by \newcite{bender2018data}. \\
\textbf{Dataset name:} MIM-GOLD-EL \\
\textbf{Dataset developer:} Steinunn Rut Friðriksdóttir, Valdimar Ágúst Eggertsson, Benedikt Geir Jóhannesson, Hjalti Daníelsson. \\
\textbf{Dataset license:} IGC-Corpus License\footnote{https://repository.clarin.is/repository/xmlui/page/license-gigaword-corpus} \\
\textbf{Link to dataset:} \url{https://repository.clarin.is/repository/xmlui/handle/20.500.12537/168} 
\subsection{Curation rationale}
As stated in Section~\ref{related} and \ref{corpus}, MIM-GOLD-EL is an extension of the previously available MIM-GOLD-NER corpus which is itself an extended version of the MIM-GOLD corpus \cite{loftsson2010developing}. MIM-GOLD-EL consists of over 21,000 mentions that have been linked to their corresponding Named Entities in Wikidata and is intended as training material for Icelandic or multilingual EL models. \\
The original MIM-GOLD corpus is intended as a gold standard for Icelandic POS-taggers. It consists of one million words of text that were tagged automatically and then manually corrected. The text was compiled from various sources as illustrated in \ref{fig.4} (MIM-GOLD, MIM-GOLD-NER and MIM-GOLD-EL all contain the exact same text with different annotations) which should ensure that it is generalizable and not limited to a specific domain. \\ 
As noted in Footnote \ref{foot8}, one of the files from the original MIM-GOLD corpus contains adjudications where entities representing people and locations have been anonymized. This file is only included in MIM-GOLD-EL to respect the original schema and is not well suitable for EL tasks. 

\subsection{Language variety}
The language of this corpus is Icelandic. No dialect specifications apply to Icelandic.

\subsection{Speaker demographic}
Due to the nature of the corpus, it's hard to give detailed information regarding the demographics of the authors. We can infer that all of the text were written by Icelandic authors aged between 18-70. 

\subsection{Annotator demographic}
The four annotators that worked on MIM-GOLD-EL were all Icelandic, aged between 28-42. One of the annotators has a degree in Icelandic and three are computer scientists. All of the annotators have extensive professional proficiency in computational linguistics. 

\subsection{Speech situation}
The corpus is divided into several subsections that include various types of language registers. The language used in the texts compiled from news articles is not the same as that of the blogs despite Icelandic having no dialects. However, all of the text presented in the corpus is written text (as opposed to transcriptions of speech). 

\subsection{Text characteristics}
Same applies here as with the speech situation. The text presented in the corpus is compiled from various sources and thus contains various characteristics. 

\subsection{Recording quality}
N/A

\subsection{Other}
N/A

\subsection{Provenance appendix}
See Section~\ref{related} and \ref{corpus}.

\end{document}